\documentclass[10pt,twocolumn,letterpaper]{article}

\usepackage{cvpr}
\usepackage{times}
\usepackage{epsfig}
\usepackage{graphicx}
\usepackage{amsmath}
\usepackage{amssymb}
\usepackage{dsfont}
\usepackage{gensymb}
\usepackage{comment}
\usepackage{subfigure}
\usepackage{pifont}
\usepackage{multirow}
\usepackage[multiple]{footmisc}
\usepackage{tabularx,ragged2e,booktabs}
\usepackage{pbox}

\usepackage[pagebackref=true,breaklinks=true,letterpaper=true,colorlinks,bookmarks=false]{hyperref}

\cvprfinalcopy % *** Uncomment this line for the final submission

\def\cvprPaperID{205} % *** Enter the CVPR Paper ID here

\def\tabletext{\small}
\newcommand{\specialcell}[2][c]{%
  \begin{tabular}[#1]{@{}c@{}}#2\end{tabular}}
\newcommand{\cmark}{\ding{51}}%

\begin{document}

%%%%%%%%% TITLE

\title{EgoSampling: Fast-Forward and Stereo for Egocentric Videos}

\author{Yair Poleg\\
The Hebrew University\\
Jerusalem, Israel\\
\and
Tavi Halperin\\
The Hebrew University\\
Jerusalem, Israel\\
\and
Chetan Arora\\
IIIT\\
Delhi, India\\
\and
Shmuel Peleg\\
The Hebrew University\\
Jerusalem, Israel
}

\maketitle
%\thispagestyle{empty}

%%%%%%%%% ABSTRACT
\begin{abstract}

    While egocentric cameras like GoPro are gaining popularity, the videos they capture are long, boring, and difficult to watch from start to end. Fast forwarding (i.e. frame sampling) is a natural choice for faster video browsing. However, this accentuates the shake caused by natural head motion, making the fast forwarded video useless.

    We propose EgoSampling, an adaptive frame sampling that gives more stable fast forwarded videos. Adaptive frame sampling is formulated as energy minimization, whose optimal solution can be found in polynomial time.

    In addition, egocentric video taken while walking suffers from the left-right movement of the head as the body weight shifts from one leg to another. We turn this drawback into a feature: Stereo video can be created by sampling the frames from the left most and right most head positions of each step, forming approximate stereo-pairs.
\end{abstract}

\section{Introduction}

With the increasing popularity of GoPro \cite{gopro} and the introduction of Google Glass \cite{glass} the use of head worn egocentric cameras is on the rise. These cameras are typically operated in a hands-free, always-on manner, allowing the wearers to concentrate on their activities. While more and more egocentric videos are being recorded, watching such videos from start to end is difficult due to two aspects: (i) The videos tend to be long and boring; (ii) Camera shake induced by natural head motion further disturbs viewing. These aspects call for automated tools to enable faster access to the information in such videos.
An exceptional tool for this purpose is the ``Hyperlapse" method recently proposed by \cite{hyperlapse}. While our work was inspired by \cite{hyperlapse}, we take a different, lighter, approach to address this problem.

\begin{figure}[t]
    \centering
    %\subfigure[]{\includegraphics[width=1\columnwidth]{figures/fig-schematic-ff1-allframes.pdf}}
    \subfigure[]{\includegraphics[width=0.9\columnwidth]{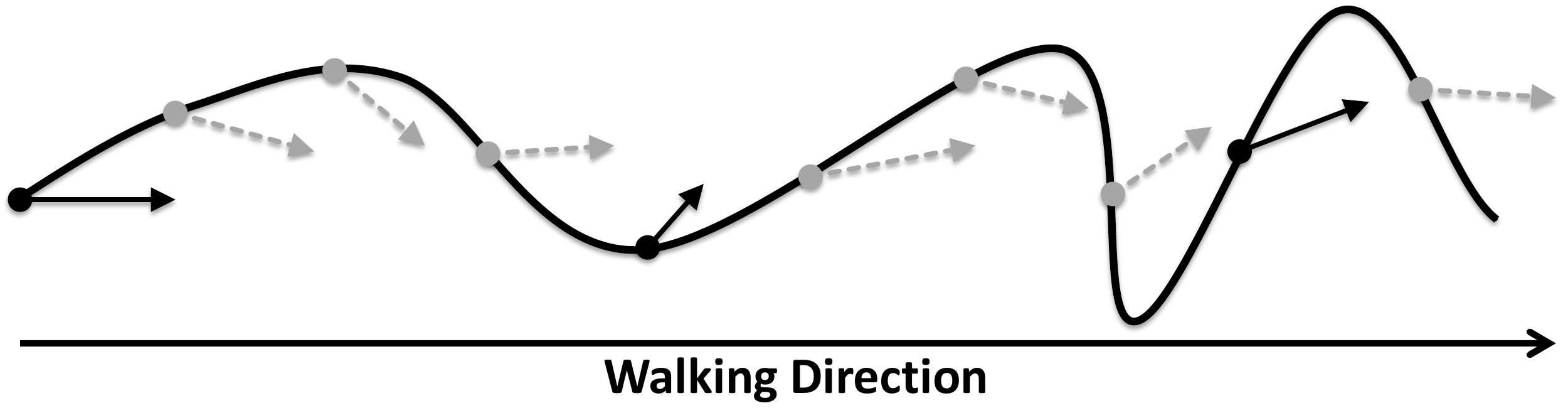}} \\
    \subfigure[]{\includegraphics[width=0.9\columnwidth]{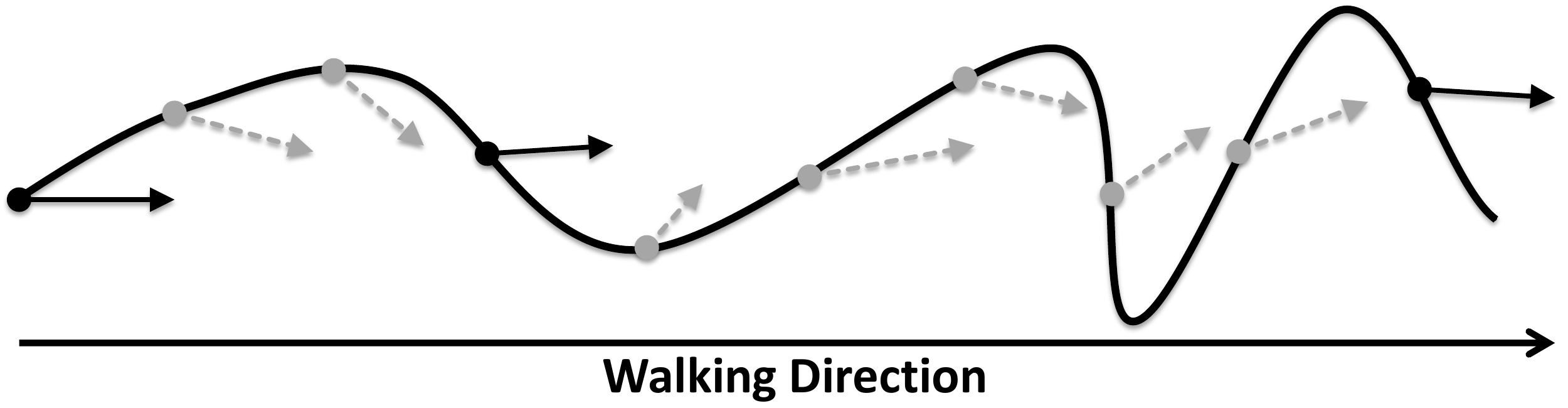}}
    \caption{Frame sampling for Fast Forward. A view from above on the camera path (the line) and the viewing directions of the frames (the arrows) as the camera wearer walks forward during a couple of seconds. (a) Uniform $5\times$ frames sampling, shown with solid arrows, gives output with significant changes in viewing directions. (b) Our frame sampling, represented as solid arrows, prefers forward looking frames at the cost of somewhat non uniform sampling.}
    \label{fig:ff-schematic}
\end{figure}

Fast forward is a natural choice for faster browsing of egocentric videos. The speed factor depends on the cognitive load a user is interested in taking. Na\"{\i}ve fast forward uses uniform sampling of frames, and the sampling density depends on the desired speed up factor. Adaptive fast forward approaches \cite{Petrovic:2005} try to adjust the speed in different segments of the input video so as to equalize the cognitive load. For example, sparser frame sampling giving higher speed up is possible in stationary scenes, and denser frame sampling giving lower speed ups is possible in dynamic scenes. In general, content aware techniques adjust the frame sampling rate based upon the importance of the content in the video. Typical importance measures include scene motion, scene complexity, and saliency. None of the aforementioned methods, however, can handle the challenges of egocentric videos, as we describe next.

\begin{figure*}[ht]
	\centering
	\includegraphics[width=1\textwidth]{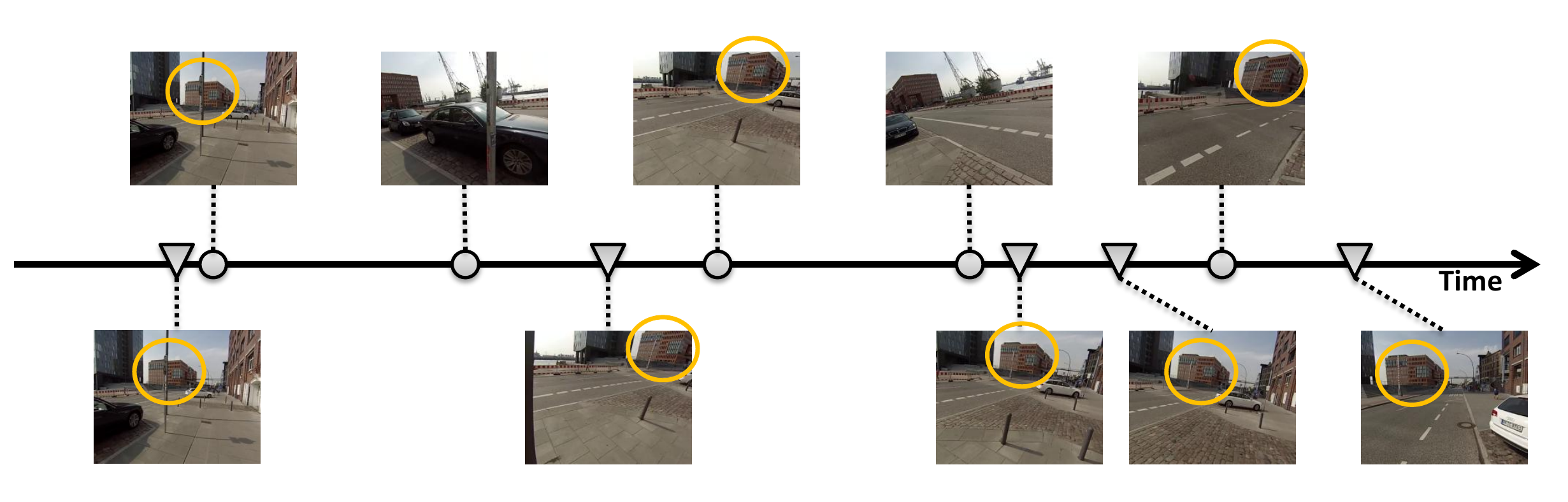}
    \caption{Representative frames from the fast forward results on `Bike2' sequence \cite{hyperlapse-dataset}. The camera wearer rides a bike and prepares to cross the road.
\underline{Top row:} uniform sampling of the input sequence leads to a very shaky output as the camera wearer turns his head sharply to the left and right before crossing the road.
\underline{Bottom row:} EgoSampling prefers forward looking frames and therefore samples the frames non-uniformly so as to remove the sharp head motions. The stabilization can be visually compared by focusing on the change in position of the building (circled yellow) appearing in the scene. The building does not even show up in two frames of the uniform sampling approach, indicating the extreme shake. Note that the fast forward sequence produced by EgoSampling can be post-processed by traditional video stabilization techniques to further improve the stabilization.}
    \label{fig:result_bike2}
\end{figure*}

Most egocentric videos suffer from substantial camera shake due to natural head motion of the wearer. We borrow the terminology of \cite{us} and note that when the camera wearer is ``stationary" (e.g, sitting or standing in place), head motions are less frequent and pose no challenge to traditional fast-forward and stabilization techniques. However, when the camera wearer is ``in transit" (e.g, walking, cycling, driving, etc), existing fast forward techniques end up accentuating the shake in the video. We therefore focus on handling these cases, leaving the simpler cases of a stationary camera wearer for standard methods. We use the method of \cite{us} to identify with high probability portions of the video in which the camera wearer is not ``stationary", and operate only on these. Other methods, such as \cite{kitani,grauman-story} can also be used to identify a stationary camera wearer.

We propose to model frame sampling as an energy minimization problem. A video is represented as a directed a-cyclic graph whose nodes correspond to input video frames. The weight of an edge between nodes, e.g. between frame $t$ and frame $t+k$, represents a cost for the transition from $t$ to $t+k$. For fast forward, the cost represents how ``stable" the output video will be if frame $t$ is followed by frame $t+k$ in the output video. This can also be viewed as introducing a bias favoring a smoother camera path. The weight will additionally indicate how suitable $k$ is to the desired playback speed. In this formulation, the problem of generating a stable fast forwarded video becomes equivalent to that of finding a shortest path in a graph. We keep all edge weights non-negative and note that there are numerous, polynomial time, optimal inference algorithms available for finding a shortest path in such graphs. We show that sequences produced with our method are more stable and easier to watch compared to traditional fast forward methods.

An interesting phenomenon of a walking person is the shifting of body weight from one leg to the other leg, causing periodic head motion from left to right and back. Given an egocentric video taken by a walking person, sampling  frames from the left most and right most head positions gives approximate stereo-pairs. This enables  generation of a stereo video from a monocular input video.

The contributions of this papers are: (i) A novel and lightweight approach for creating fast forward videos for egocentric videos. (ii) A method to create stereo sequences from monocular egocentric video.

The rest of this paper is organized as follows. We survey related works in Section \ref{sec:related_work}. Proposed frame sampling method for fast forward and problem formulation are presented in Sections \ref{sec:sampling_framework} and \ref{sec:problem_formulation} respectively. In Section \ref{sec:stereo} we describe our method for creating
perceptual stereo sequences. Experiments and user study results are given in Section \ref{sec:exp}. We conclude in Section \ref{sec:concl}.

\section{Related Work}
\label{sec:related_work}

\paragraph*{Video Summarization:}

Video Summarization methods sample the input video for salient events to create a concise output that captures the essence of the input video. This field has seen many new papers in the recent years, but only  a handful address the specific challenges of summarizing egocentric videos. In \cite{grauman-important-people,grauman-snap-points}, important keyframes are sampled from the input video to create a story-board summarizing the input video. In \cite{grauman-story}, subshots that are related to the same ``story" are sampled to produce a ``story-driven" summary. Such video summarization can be seen as an extreme adaptive fast forward, where some parts are completely removed while other parts are played at original speed. These techniques are required to have some strategy for determining the importance or relevance of each video segment, as segments removed from summary are not available for browsing. As long as automatic methods are not endowed with human intelligence, fast forward gives a person the ability to survey all parts of the video.

\paragraph*{Video Stabilization:}

There are two main approaches for video stabilization. One approach uses $3D$ methods to reconstruct a smooth camera path \cite{content_preserving_warps,vid3d_stab_depth}. Another approach avoids $3D$, and uses only $2D$ motion models followed by non-rigid warps \cite{youtube_stabilizer,subspace_vid_stab,BundledPaths2013,steadyflow,raanan}. A na\"{\i}ve fast forward approach would be to apply video stabilization algorithms before or after uniform frame sampling. As noted by \cite{hyperlapse} also, stabilizing egocentric video doesn't produce satisfying results. This can be attributed to the fact that uniform sampling, irrespective of whether done before or after the stabilization, is not able to remove outlier frames, e.g. the frames when camera wearer looks at his shoe for a second while walking in general.

An alternative approach that was evaluated in \cite{hyperlapse}, termed ``coarse-to-fine stabilization", stabilizes the input video and then prunes frames from the stabilized video a bit. This process is repeated until the desired playback speed is achieved. Being a uniform sampling approach, this method does not avoid outlier frames. In addition, it introduces significant distortion to the output as a result of repeated application of a stabilization algorithm.

EgoSampling differs from traditional fast forward as well as traditional video stabilization. We attempt to adjust frame sampling in order to produce a stable-as-possible fast forward sequence. Rather than stabilizing outlier frames, we prefer to skip them. While traditional stabilization algorithms must make compromises (in terms of camera motion and crop window) in order to deal with every outlier frame, we have the benefit of choosing which frames to include in the output. Following our frame sampling, traditional video stabilization algorithms \cite{youtube_stabilizer,subspace_vid_stab,BundledPaths2013,steadyflow,raanan} can be applied to the output of EgoSampling to further stabilize the results.

\paragraph*{Hyperlapse:}

A recent work \cite{hyperlapse}, dedicated to egocentric videos, proposed to use a combination of $3D$ scene reconstruction and image based rendering techniques to produce a completely new video sequence, in which the camera path is perfectly smooth and broadly follows the original path. The results of Hyperlapse are impressive. However, the scene reconstruction and image based rendering methods are not guaranteed to work for many egocentric videos, and the computation costs involved are very high. Hyperlapse may therefore be less practical for day-long videos which need to be processed at home. Unlike Hyperlapse, EgoSampling uses only raw frames sampled from the original video.

\section{Proposed Frame Sampling}
\label{sec:sampling_framework}

Most egocentric cameras are usually worn on the head or attached to eyeglasses. While this gives an ideal first person view, it also leads to significant shaking of the camera due to the wearer's head motion. Camera Shaking is higher when the person is ``in transit" (e.g. walking, cycling, driving, etc.). In spite of the shaky original video, we would prefer for consecutive output frames in the fast forward video to have similar viewing directions, almost as if they were captured by a camera moving forward on rails. In this paper we propose a frame sampling technique, which selectively picks frames with similar viewing directions, resulting in a stabilized fast forward egocentric video. See Fig.~\ref{fig:ff-schematic} for a schematic example.

\paragraph*{Head Motion Prior}

As noted by \cite{us,kitani,grauman-important-people,ryoo}, the camera shake in an egocentric video, measured as optical flow between two consecutive frames, is far from being random. It contains enough information to recognize the camera wearer's activity. Another observation made in \cite{us} is that when ``in transit'', the mean (over time) of the instantaneous optical flow is always radially away from the Focus of Expansion (FOE). The interpretation is simple: when ``in transit'' (e.g., walking/cycling/driving etc), our head might be moving instantaneously in all directions (left/right/up/down), but the physical transition between the different locations is done through the forward looking direction (i.e. we look forward and move forward). This motivates us to use a forward orientation sampling prior. When sampling frames for fast forward, we prefer frames looking to the direction in which the camera is translating.

\paragraph*{Computation of Motion Direction (Epipole)}

Given $N$ video frames, we would like to find the motion direction (Epipolar point) between all pairs of frames, $I_t$ and $I_{t+k}$, where $k \in [1,\tau]$, and $\tau$ is the maximum allowed frame skip. Under the assumption that the camera is always translating (when the camera wearer is ``in transit''), the displacement direction between $I_t$ and $I_{t+k}$ can be estimated from the fundamental matrix $F_{t,t+k}$ \cite{hartley_book}. Frame sampling will be biased towards selecting forward looking frames, where the epipole is closest to the center of the image.
Recent V-SLAM approaches such as \cite{lsdslam,svo} provide camera ego-motion estimation and localization in real-time. However, these methods failed on our dataset after a few hundreds frames. We decided to stick with robust $2D$ motion models.

\begin{comment}
\begin{figure}[t]
    \centering
    \includegraphics[width=0.8\columnwidth]{figures/fig_fmat_decay.pdf}
    \caption{The decrease in successful computation of the Fundamental Matrix as the two frames become farther apart temporally. The x-axis is the number of skipped frames between the two frames used for computation.}
	\label{fig:fmat_decay}
\end{figure}
\end{comment}

\paragraph*{Estimation of Motion Direction (FOE)}

We found that the fundamental matrix computation can fail frequently when $k$ (temporal separation between the frame pair) grows larger.
%Figure \ref{fig:fmat_decay} shows the fast decay in the number of fundamental matrices computed successfully as a function of $k$.
Whenever the fundamental matrix computation breaks, we estimate the direction of motion from the FOE of the optical flow. We do not compute the FOE from the instantaneous flow, but from integrated optical flow as suggested in \cite{us} and computed as follows: (i) We first compute the sparse optical flow between all consecutive frames from frame $t$ to frame $t+k$. Let the optical flow between frames $t$ and $t+1$ be denoted by $g_t(x,y)$. (ii) For each flow location $(x,y)$, we average all optical flow vectors at that location from all consecutive frames. $G(x,y) =  \frac{1}{k} \sum_{i=t}^{t+k-1} g_i(x,y)$. The FOE is computed from $G$ according to \cite{technion-foe}, and is used as an estimate of the direction of motion.

The temporal average of optical flow gives a more accurate FOE since the direction of translation is relatively constant, but the head rotation goes to all directions, back and forth. Averaging the optical flow will tend to cancel the rotational components, and leave the translational components. In this case the FOE is a good estimate for the direction of motion. For a deeper analysis of temporally integrated optical flow see ``Pixel Profiles'' in \cite{steadyflow}.

\paragraph*{Optical Flow Computation}

Most available algorithms for dense optical flow failed for our purposes, but the very sparse flow proposed in \cite{us} for egocentric videos worked relatively well. The fifty optical flow vectors were robust to compute, while allowing to find the FOE quite accurately.

\section{Problem Formulation and Inference}
\label{sec:problem_formulation}
\begin{figure}[t]
    \centering
    \includegraphics[width=1\linewidth]{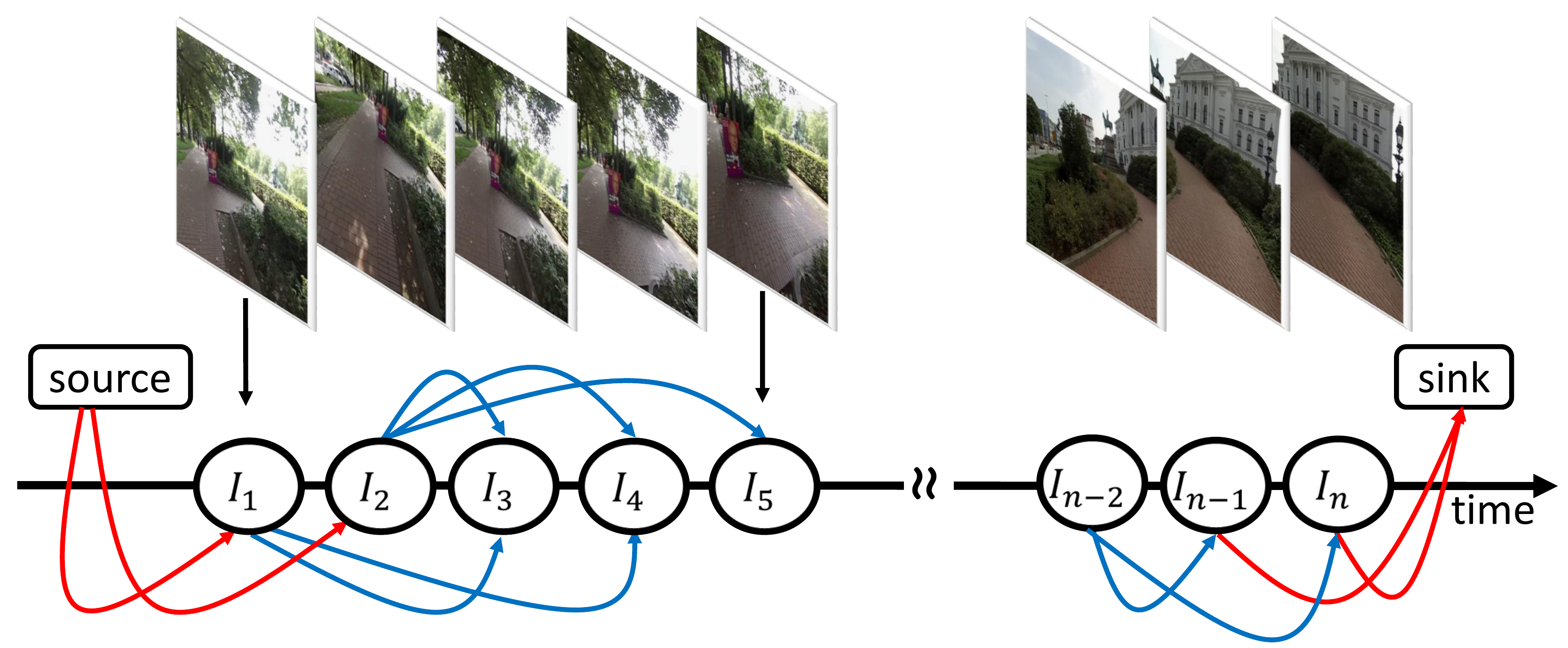}
    \caption{We formulate the joint fast forward and video stabilization problem as finding a shortest path in a graph constructed as shown. There is a node corresponding to each frame. The edges between a pair of frames $(i,j)$ indicate the penalty for including a frame $j$ immediately after frame $i$ in the output (please refer to the text for details on the edge weights). The edges between source/sink and the graph nodes allow to skip frames from start and end. The frames corresponding to nodes along the shortest path from source to sink are included in the output video.}
    \label{fig:first_order_graph}
\end{figure}

\begin{figure*}[t]
    \centering
	\includegraphics[width=0.8\linewidth]{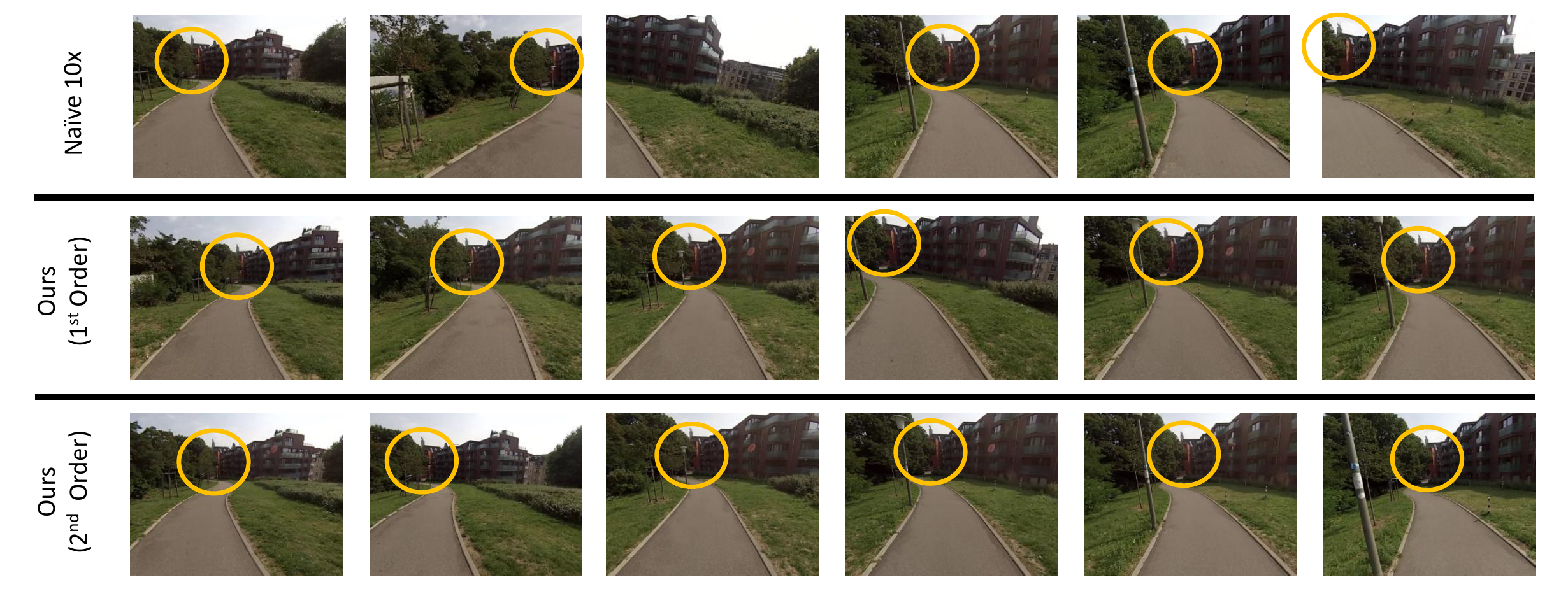}
    \caption{Comparative results for fast forward from na\"{\i}ve uniform sampling (first row), EgoSampling using first order formulation (second row) and using second order formulation (third row). Note the stability in the sampled frames as seen from the tower visible far away (circled yellow). The first order formulation leads to a more stable fast forward output compared to na\"{\i}ve uniform sampling. The second order formulation produces even better results in terms of visual stability.}
    \label{fig:res_ff_comparison}
\end{figure*}

We model the joint fast forward and stabilization of egocentric video as an energy minimization problem. We represent the input video as a graph with a node corresponding to every frame in the video. There are weighted edges between every pair of graph nodes, $i$ and $j$, with weight proportional to our preference for including frame $j$ right after $i$ in the output video. There are three components in this weight:

\begin{enumerate}
    \item Shakiness Cost ($S_{i,j}$): This term prefers forward looking frames. The cost is proportional to the distance of the computed motion direction (Epipole or FOE) from the center of the image.
    \item Velocity Cost ($V_{i,j}$): This term controls the playback speed of the output video. The desired speed is given by the desired magnitude of the optical flow, $K_{flow}$, between two consecutive output frames. This optical flow is estimated as follows: (i) We first compute the sparse optical flow between all consecutive frames from frame $i$ to frame $j$. Let the optical flow between frames $t$ and $t+1$ be $g_t(x,y)$. (ii) For each flow location $(x,y)$, we sum all optical flow vectors at that location from all consecutive frames. $G(x,y) = \sum_{t=i}^{j-1} g_t(x,y)$.  (iii) The flow between frames $i$ and $j$ is then estimated as the average magnitude of all the flow vectors $G(x,y)$. The closer the magnitude is to $K_{flow}$, the lower is the velocity cost.

        The velocity term samples more densely periods with fast camera motion compared to periods with slower motion, e.g. it will prefer to skip stationary periods, such as when waiting at a red light. The term additionally brings in the benefit of content aware fast forwarding. When the background is close to the wearer, the scene changes faster compared to when the background is far away. The velocity term reduces the playback speed when the background is close and increases it when the background is far away.
    \item Appearance Cost ($C_{i,j}$): This is the Earth Movers Distance (EMD) \cite{emd} between the color histograms of frames $i$ and $j$. The role of this term is to prevent large visual changes between frames. A quick rotation of the head or dominant moving objects in the scene can confuse the FOE or epipole computation. The terms acts as an anchor in such cases, preventing the algorithm from skipping a large number of frames.
\end{enumerate}

The overall weight of the edge between nodes (frames) $i$ and $j$ is given by:
\begin{equation}
\mathcal{W}_{i,j}=\alpha\cdot\mathcal{S}_{i,j}+\beta\cdot V_{i,j}+\gamma\cdot C_{i,j},
\end{equation}
where $\alpha$, $\beta$ and $\gamma$ represent the relative importance of various costs in the overall edge weight.

With the problem formulated as above, sampling frames for stable fast forward is done by finding a shortest path in the graph. We add two auxiliary nodes, a \emph{source} and a \emph{sink} in the graph to allow skipping some frames from start or end. We add zero weight edges from start node to first $D_{start}$ frames and from last $D_{end}$ nodes to sink, to allow such skip. We then use Dijkstra's algorithm \cite{dijkstra} to compute the shortest path between source and sink. The algorithm does the optimal inference in time polynomial in the number of nodes (frames). Fig.~\ref{fig:first_order_graph} shows a schematic illustration of the proposed formulation.

We note that there are content aware fast forward and other general video summarization techniques which also measure importance of a particular frame being included in the output video, e.g. based upon visible faces or other objects. In our implementation we have not used any bias for choosing a particular frame in the output video based upon such a relevance measure. However, the same could have been included easily. For example, if the penalty of including a frame, $i$, in the output video is $\delta_i$, the weights of all the incoming (or outgoing, but not both) edges to node $i$ may be increased by $\delta_i$.

\subsection{Second Order Smoothness}

\begin{figure}[t]
    \centering
    \includegraphics[width=1\linewidth]{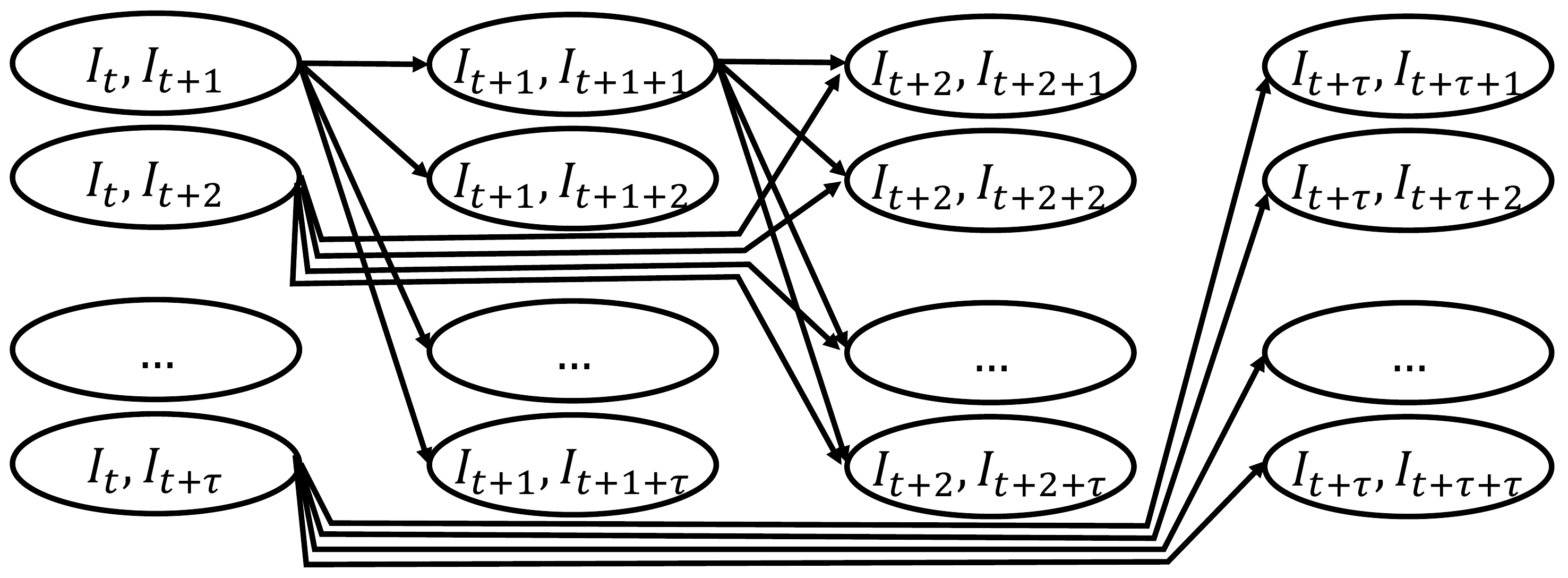}
    \caption{The graph formulation, as described in Fig.~\ref{fig:first_order_graph}, produces an output which has almost forward looking direction. However, there may still be large changes in the epipole locations between two consecutive frame transitions, causing jitter in the output video.
To overcome this we add a second order smoothness term based on triplets of output frames.
    Now the nodes correspond to pair of frames, instead of single frame in first order formulation described earlier. There are edges between frame pairs $(i,j)$ and $(k,l)$,  if $j=k$. The edge reflects the penalty for including frame triplet $(i,k,l)$ in the output. Edges from source and sink to graph nodes (not shown in the figure) are added in the same way as in first order formulation to allow skipping frames from start and end.}
    \label{fig:second_order_graph}
\end{figure}

The formulation described in the previous section prefers to select forward looking frames, where the epipole is closest to the center of the image. With the proposed formulation, it may so happen that the epipoles of the selected frames are close to the image center but on the opposite sides, leading to a jitter in the output video. In this section we introduce an additional cost element: stability of the location of the epipole. We prefer to sample frames with minimal variation of the epipole location.

To compute this cost, nodes now represent two frames, as can be seen in Fig.~\ref{fig:second_order_graph}. The weights on the edges depend on the change in epipole location between one image pair to the successive image pair. Consider three frames $I_{t_1}$, $I_{t_2}$ and $I_{t_3}$. Assume the epipole between $I_{t_i}$ and $I_{t_j}$ is at pixel $(x_{ij}, y_{ij})$. The second order cost of the triplet (graph edge)  $(I_{t_1},I_{t_2},I_{t_3})$, is proportional to $\|(x_{23}-x_{12}, y_{23}-y_{12})\|$. This is the difference between the epiople location computed from frames $I_{t_1}$ and $I_{t_2}$, and the epipole location computed from frames $I_{t_2}$ and $I_{t_3}$.

This second order cost is added to the previously computed shakiness cost, which is proportional to the distance from the origin $\|(x_{23}, y_{23})\|$. The graph with the second order smoothness term has all edge weights non-negative and the running-time to find optimal solution to shortest path is linear in the number of nodes and edges, i.e. $O(n\tau^2)$. In practice, with $\tau=100$, the optimal path was found in all examples in less than 30 seconds. Fig.~\ref{fig:res_ff_comparison} shows results obtained from both first order and second order formulations.

As noted for the first order formulation, we do not use importance measure for a particular frame being added in the output in our implementation. To add such, say for frame $i$, the weights of all incoming (or outgoing but not both) edges to all nodes $(i,j)$ may be increased by $\delta_i$, where $\delta_i$ is the penalty for including frame $i$ in the output video.

\section{Turning Egocentric Video to Stereo} \label{sec:stereo}

\begin{figure}[t]
    \centering
    \includegraphics[width=0.8\columnwidth]{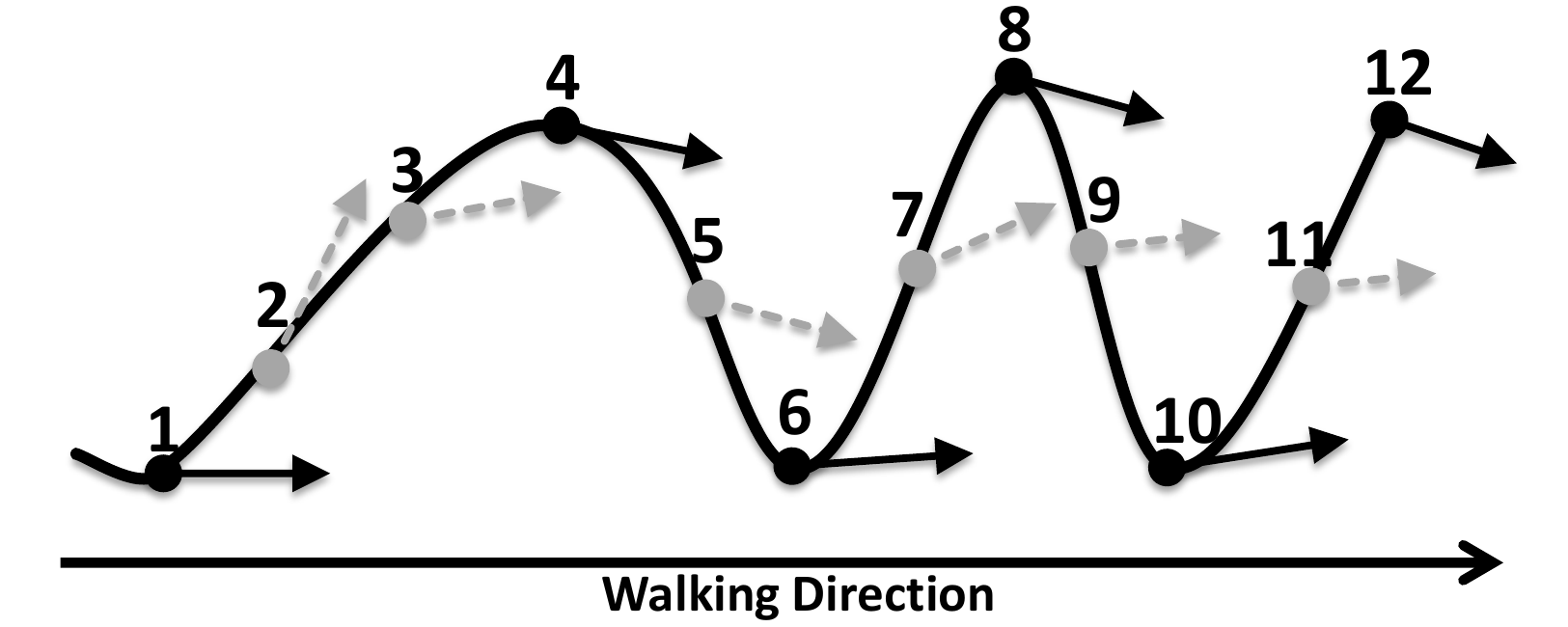}
    \caption{Frame sampling for Stereo: A view from above for the camera path (the line) and the viewing directions of the frames (numbered arrows). The camera wearer walks forward for a couple of seconds. We pick the frames in which the wearer's head is in the right most position (frames 1,6,10) and left most position (frames 4,8,12) to form stereo pairs. Frame pairs (1,4), (6,8) and (10,12) form the output stereo video.}
    \label{fig:stereo-formulation}
\end{figure}

\begin{figure}[t]
    \centering
    %\subfigure{\includegraphics[height=0.2\linewidth]{figures/fig-stereo-walking4-L-frame_000301_small.jpg}}
    %\subfigure{\includegraphics[height=0.2\linewidth]{figures/fig-stereo-walking4-R-frame_000314_small.jpg}}
    \subfigure{\includegraphics[height=0.37\linewidth]{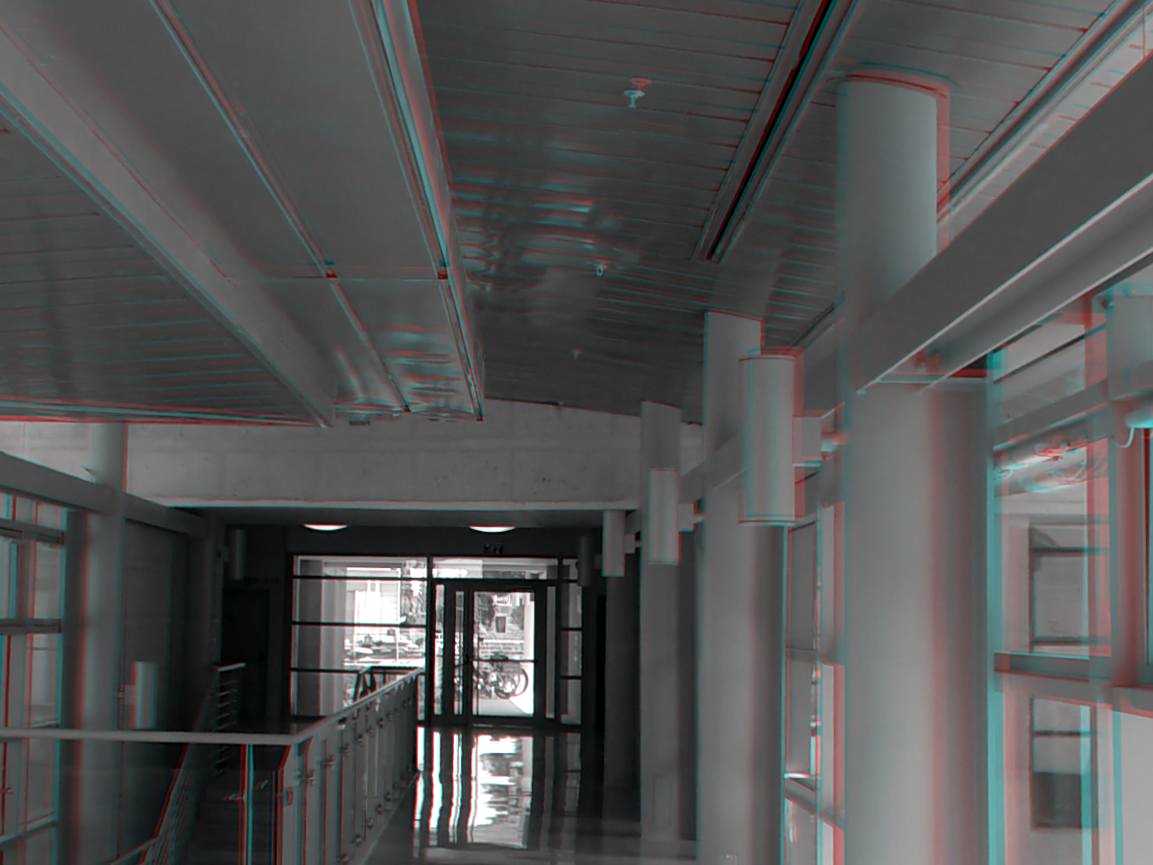}}
    \subfigure{\includegraphics[height=0.37\linewidth]{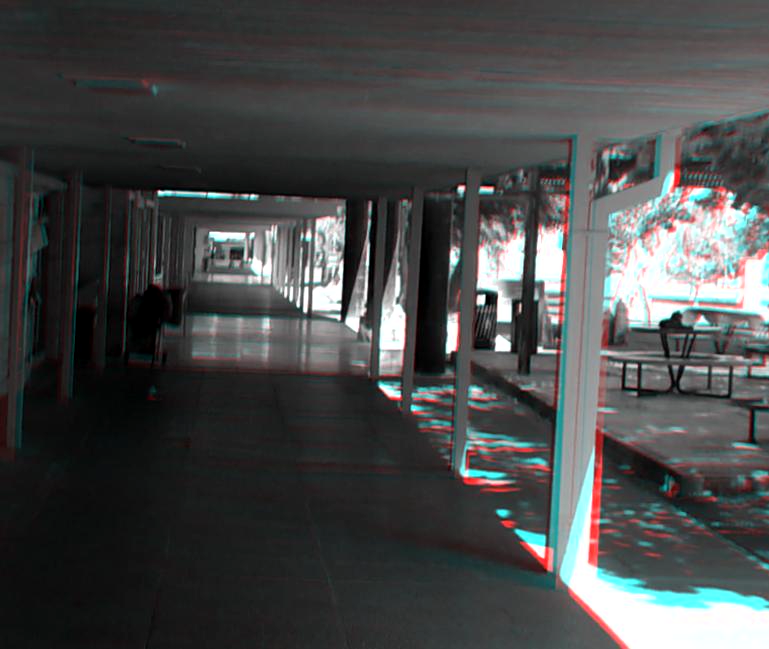}}
    \caption{Two stereo results obtained from our method. The output is shown as anaglyph composite. Please use cyan and red anglyph glasses and zoom to 800\% for best view. Readers without anaglyph glasses may note the observed disparity evident from red separation at various pixels. There is higher disparity and larger red separation on the objects near to observer. Stereo video output for these examples are available in the project page\protect\footnotemark[1].}
    \label{fig:stereo_walking4}
\end{figure}
%\footnotetext{http://www.vision.huji.ac.il/egosampling/}

When walking, the head moves left and right as the body shifts its weight from the left leg to the right leg and back. Pictures taken during the shift of the head to the left and to the right can be used to generate stereo egocentric video. For this purpose we would like to generate two stabilized videos: The left video will sample frames taken when the head moved to the left, and the right video will sample frames taken when the head moved to the right. Fig.~\ref{fig:stereo-formulation} gives the schematic approach for generating stereo egocentric videos.

For generating the stereo streams we need to determine the head location. We found the following to work well: (i) Average all optical flow vectors in each frame, and keep one scalar describing the average x-shift for that frame. (ii) Compute for each frame the accumulated x-shift of all preceding frames starting from the first frame. The curve of the accumulated x-shift is very similar to the camera path shown in Fig.~\ref{fig:stereo-formulation}. Frames near the left peaks are selected for the left video, and frames near the right peaks are selected for the right video.

In perfect stereo pairs the displacement between the two images is a pure sideways translation. In our case we also have forward motion between the two views. The forward motion can disturb stereo perception for objects which are too close, but for objects farther away stereo output produced from the proposed scheme looks good. Fig.~\ref{fig:stereo_walking4} shows frames from a stereo video generated using proposed framework.

\section{Experiments}
\label{sec:exp}

In this section we give implementation details and show the results for fast forward as well as stereo. We use publicly available sequences \cite{hyperlapse-dataset, youtube-running1, youtube-driving2, ego_social} as well as our own videos (for the stereo only) for the demonstration. We used a modified (faster) implementation of \cite{us} for the LK \cite{lk} optical flow estimation. We use the code and calibration details given by \cite{hyperlapse} to correct for lens distortion in their sequences. Feature point extraction and fundamental matrix recovery is performed using VisualSFM \cite{visualsfm}, with GPU support. The rest of the implementation (FOE estimation, energy terms and shortest path etc.) is in Matlab. All the experiments have been conducted on a standard desktop PC.

\subsection{Fast Forward}

\renewcommand{\tabcolsep}{0.1cm}
\begin{table}[t]
    \centering
    \begin{tabular}{lccccc}
        \toprule[1.5pt]
        \specialcell{\bf \tabletext Name}  & \specialcell{\bf \tabletext Src}  &\specialcell{\bf \tabletext Resolution} & \specialcell{\bf \tabletext Camera} & \specialcell{\bf \tabletext Num\\ \bf \tabletext Frames} & \specialcell{\bf \tabletext Lens \\ \bf \tabletext Correction}  \\ \midrule

        \tabletext Walking1		&	\cite{hyperlapse-dataset}	&  \specialcell{\tabletext $1280$x$960$}	& \tabletext Hero2 &	\tabletext $17249$  &	\cmark  \\        	
        \tabletext Walking2	&	\cite{hyperlapse-dataset}	&\specialcell{\tabletext $1280$x$720$}		&	\tabletext Hero	  &	\tabletext $6900$	&  	 	 \\
        \tabletext Walking3	&	\cite{us}	&  	\specialcell{\tabletext $1920$x$1080$} &	\tabletext Hero3  &	\tabletext $8001$	& 	  \\
        % In all internal docs, this experiment is called 'Driving2'.
        \tabletext Driving		&	\cite{youtube-driving2}	& 	\specialcell{\tabletext $1280$x$720$ }  		&	\tabletext Hero2	  &	\tabletext $10200$  &  \\ %
        \tabletext Bike1		&	\cite{hyperlapse-dataset}	& \specialcell{\tabletext $1280$x$960$}	&	\tabletext Hero3	  &	\tabletext $10786$  & \cmark \\
        \tabletext Bike2		&	\cite{hyperlapse-dataset}	& \specialcell{\tabletext  $1280$x$960$}	&	\tabletext Hero3	  &	\tabletext $7049$   &  	\cmark \\
        \tabletext Bike3		&	\cite{hyperlapse-dataset}	& \specialcell{\tabletext  $1280$x$960$}	&	\tabletext Hero3	  &	\tabletext $23700$  &  	\cmark \\
        % In all internal docs, this experiment is called 'Running1':
        \tabletext Running		&	\cite{youtube-running1}	& 	\specialcell{\tabletext $1280$x$720$} &	\tabletext Hero3+  &	\tabletext $12900$  &   \\

        \bottomrule[1.5pt] \\
    \end{tabular}
    \caption{Sequences used for the fast forward algorithm evaluation. All sequences were shot in 30fps, except `Running1' which is 24fps and `Walking3' which is 15fps.}
    \label{tb:ff_sequences}
\end{table}

We show results for EgoSampling on $8$ publicly available sequences. The details of the sequences are given in Table \ref{tb:ff_sequences}. For the $4$ sequences for which we have camera calibration information (marked with checks in the `Lens Correction' column), we estimated the motion direction based on epipolar geometry. We used the FOE estimation method as a fallback when we could not recover the fundamental matrix. For this set of experiments we fix the following weights: $\alpha=1000$, $\beta=200$ and $\gamma=3$. We further penalize the use of estimated FOE instead of the epipole with a constant factor $c=4$. In case camera calibration is not available, we used the FOE estimation method only and changed $\alpha=3$ and $\beta=10$. For all the experiments, we fixed $\tau=100$ (maximum allowed skip). We set the source and sink skip to $D_{start}=D_{end}=120$ to allow more flexibility. We set the desired speed up factor to $10\times$ by setting $K_{flow}$ to be $10$ times the average optical flow magnitude of the sequence. We show representative frames from the output for one such experiment in Fig.\ref{fig:res_ff_comparison}. Output videos from other experiments are given in the supplementary material\footnote[1]{http://www.vision.huji.ac.il/egosampling/}.

\paragraph*{Running times}
The advantage of the proposed approach is in its simplicity, robustness and efficiency. This makes it practical for long unstructured egocentric video. We present the coarse running time for the major steps in our algorithm below. The time is estimated on a standard Desktop PC, based on the implementation details given above. Sparse optical flow estimation (as in \cite{us}) takes 150 milliseconds per frame. Estimating F-Mat (including feature detection and matching) between frame $I_t$ and $I_{t+k}$ where $k\in[1,100]$ takes 450 milliseconds per input frame $I_t$. Calculating second-order costs takes 125 milliseconds per frame. This amounts to total of 725 milliseconds of processing per input frame. Solving for the shortest path, which is done once per sequence, takes up to 30 seconds for the longest sequence in our dataset ($\approx 24K$ frames). In all, running time is more than an order of magnitude faster than \cite{hyperlapse}.

\paragraph*{User Study}

We compare the results of EgoSampling, first and second order smoothness formulations, with na\"{\i}ve fast forward with $10\times$ speedup, implemented by sampling the input video uniformly. For EgoSampling the speed is not directly controlled but is targeted for $10\times$ speedup by setting $K_{flow}$ to be $10$ times the average optical flow magnitude of the sequence.

We conducted a user study to compare our results with the baseline methods. We sampled short clips (5-10 seconds each) from the output of the three methods at hand. We made sure the clips start and end at the same geographic location. We showed each of the 35 subjects several pairs of clips, before stabilization, chosen at random. We asked the subjects to state which of the clips is better in terms of stability and continuity. The majority ($75\%$) of the subjects preferred the output of EgoSampling with first-order shakeness term over the na\"{\i}ve baseline. On top of that, $68\%$ preferred the output of EgoSampling using second-order shakeness term over the output using first-order shakeness term.

To evaluate the effect of video stabilization on the EgoSampling output, we tested three commercial video stabilization tools: (i) Adobe Warp Stabilizer (ii) Deshaker \footnote[2]{http://www.guthspot.se/video/deshaker.htm} (iii) Youtube's Video stabilizer. We have found that  Youtube's stabilizer gives the best results on challenging fast forward videos \footnote[3]{We attribute this to the fact that Youtube's stabilizer does not depend upon long feature trajectories, which are scarce in sub-sampled video as ours.}. We stabilized the output clips using Youtube's stabilizer and asked our 35 subjects to repeat process described above. Again, the subjects favored the output of EgoSampling.

\paragraph*{Quantitative Evaluation}

\renewcommand{\tabcolsep}{0.1cm}
\begin{table}
    \centering
    \begin{tabular}{lccccccc}
        \toprule[1.5pt]
        \specialcell{\bf \tabletext Name}  &\specialcell{\bf \tabletext Input\\ \bf \tabletext Frames} &\specialcell{\bf \tabletext Output \\ \bf \tabletext Frames}  & \specialcell{\bf \tabletext Median \\ \bf \tabletext Skip} & \specialcell{\bf \tabletext Improvement over \\ \bf \tabletext Na\"{\i}ve $10\times$} \\  \midrule

        \tabletext Walking1		& \tabletext $17249$  &	  \tabletext $931$ &   \tabletext $17$  &  $283\%$ \\
        \tabletext Walking2	&	\tabletext $6900$	&    \tabletext $284$ &    \tabletext $13$ &	$88\%$ \\ 	
        \tabletext Walking3	& 	\tabletext $8001$	& 	 \tabletext $956$ &    \tabletext $4$ & $56\%$	 \\
        % In all internal docs, this experiment is called 'Driving2'.
        \tabletext Driving		&	\tabletext $10200$  & \tabletext $188$ &   \tabletext $48$ & $-7\%$ \\ %
        \tabletext Bike1		&	\tabletext $10786$   &  \tabletext $378$ &  \tabletext $13$ & $235\%$ \\ %  \\
        \tabletext Bike2 &	\tabletext $7049$      &  \tabletext $343$ &    \tabletext $14$	& $126\%$  \\
        \tabletext Bike3		&    \tabletext $23700$  &  \tabletext $1255$ &  \tabletext $12$ & $66\%$ \\
        % In all internal docs, this experiment is called 'Running1':
        \tabletext Running		&	\tabletext $12900$   &  \tabletext $1251$ & \tabletext $8$   & $200\%$ \\

        \bottomrule[1.5pt] \\
    \end{tabular}
    \caption{Fast forward results with desired speedup of factor $10$ using second-order smoothness. We evaluate the improvement as degree of epipole smoothness in the output video (column $5$). Please refer to text for details on how we quantify smoothness. The proposed method gives huge improvement over na\"{\i}ve fast forward in all but one test sequence (`Driving'), see Fig.~\ref{fig:ff-failure-driving} for details. Note that one of the weaknesses of the proposed method is lack of direct control over speedup factor. Though the desired speedup factor is $10$, the actual frame skip (column $4$) differs a lot from target due to conflicting constraint posed by stabilization.}
    \label{tb:ff}
\end{table}

We quantify the performance of EgoSampling using the following measures. We measure the deviation of the output from the desired speedup. We found that measuring the speedup by taking the ratio between the number of input and output frames is misleading, because one of the features EgoSampling is to take large skips when the magnitude of optical flow is rather low. We therefore measure the effective speedup as the median frame skip.

Additional measure is the reduction in epipole jitter between consecutive output frames (or FOE if F-Matrix cannot be estimated). We differentiate the locations of the epipole (temporally). The mean magnitude of the derivative gives us the amount of jitter between consecutive frames in the output. We measure the jitter for our method as well for naive $10\times$ uniform sampling and calculate the percentage improvement in jitter over competition.

Table \ref{tb:ff} shows the quantitative results for frame skip and epipole smoothness. There is a huge improvement in jitter by our algorithm. We note that the standard method to quantify video stabilization algorithms is to measure crop and distortion ratios. However since we jointly model fast forward and stabilization such measures are not applicable. The other method could have been to post process the output video with a standard video stabilization algorithm and measure these factors. Better measures might indicate better input to stabilization or better output from preceding sampling. However, most stabilization algorithms rely on trajectories and fail on resampled video with large view difference. The only successful algorithm was Youtube's stabilizer but it did not give us these measures.

\paragraph*{Limitations}

\begin{figure}[t]
    \centering
    \includegraphics[width=0.4\columnwidth]{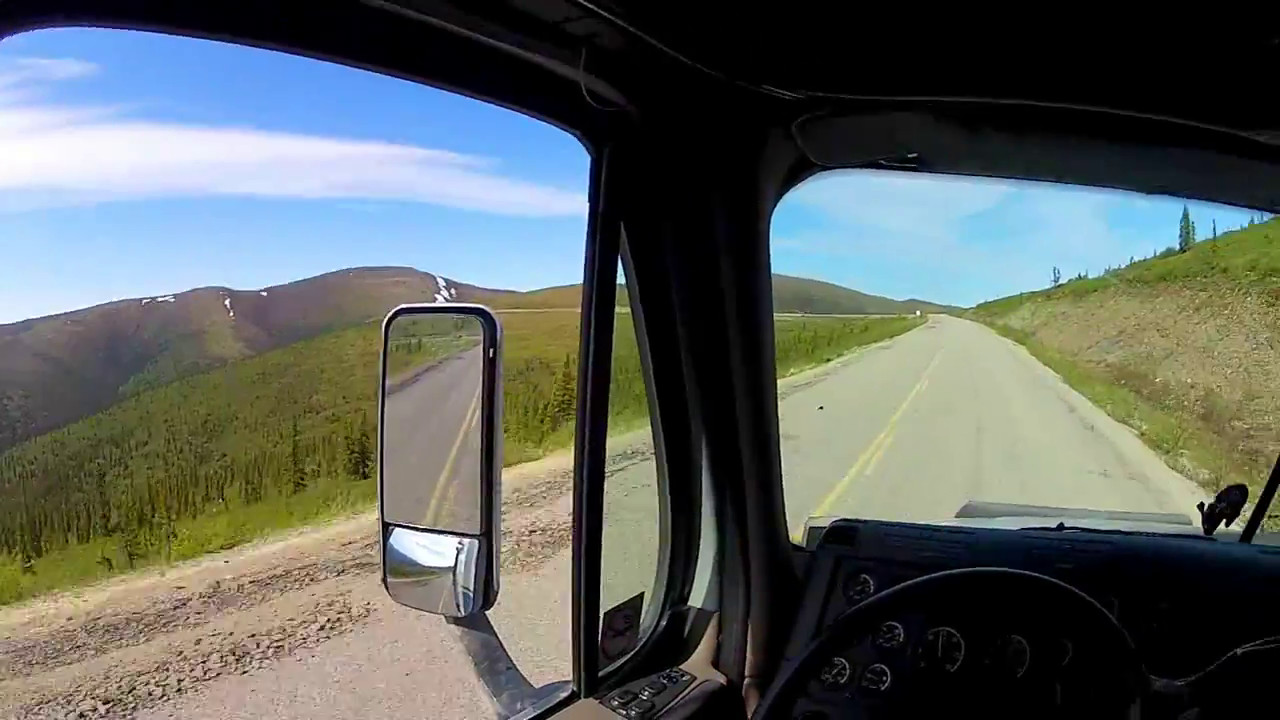}
    \includegraphics[width=0.4\columnwidth]{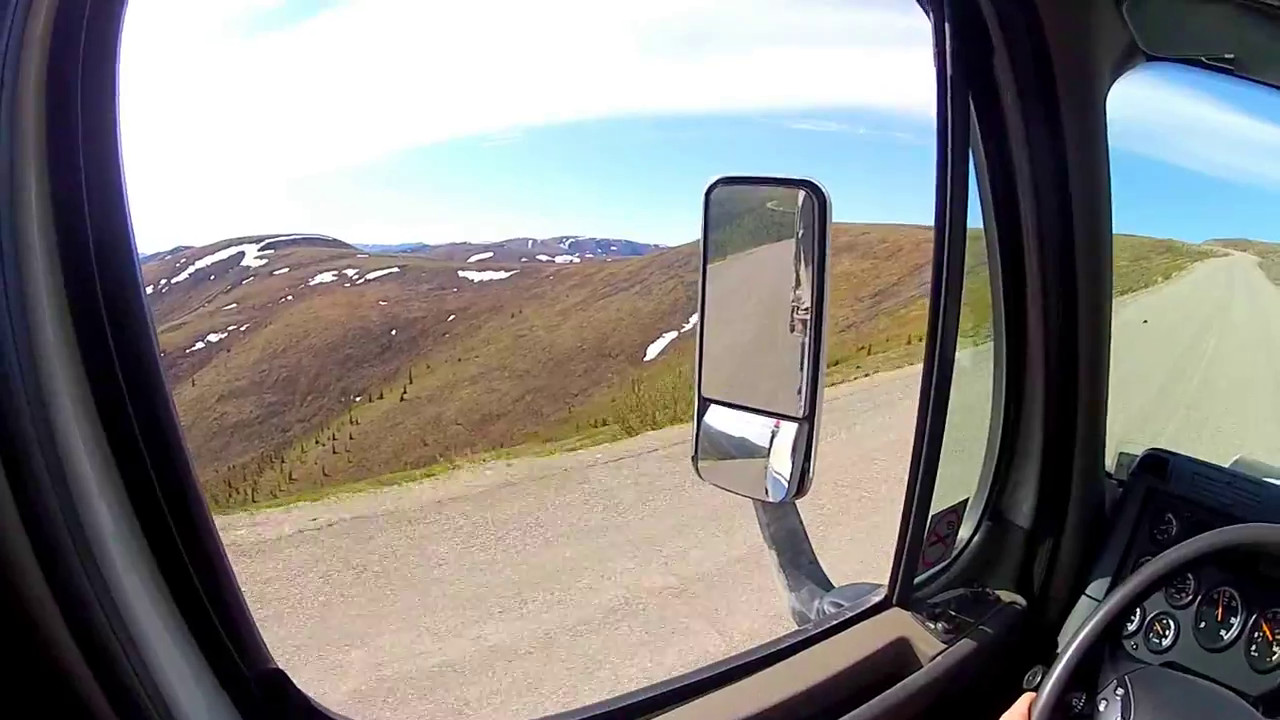}
    \caption{A failure case for the proposed method. Two sample frames from the sequence. Note that the frame to frame optical flow computed for this sequence is misleading - most of the field of view is either far away (infinity) or inside the car. In both cases, its near zero. However, since the driver shakes his head every few seconds, the average optical flow magnitude is relatively high. The velocity term causes us to skip many frames until the desired $K_{flow}$ is met, causing large frame skips in the output video. Restricting the maximum frame skip by setting $\tau$ to a small value leads to arbitrary frames being chosen looking sideways, causing shake in the output video.}
    \label{fig:ff-failure-driving}
\end{figure}

One notable difference between EgoSampling and traditional fast forward methods is that the number of output frames is not fixed. To adjust the effective speedup, the user can tune the velocity term by setting different values to $K_{flow}$. It should be noted, however, that not all speedup factors are possible without compromising the stability of the output. For example, consider a camera that toggles between looking straight and looking to the left every $10$ frames. Clearly, any speedup factor that is not a multiple of $10$ will introduce shake to the output. The algorithm chooses an optimal speedup factor which balances between the desired speedup and what can be achieved in practice on the specific input. Sequence `Driving' (Figure \ref{fig:ff-failure-driving}) presents an interesting failure case.

Another limitation of EgoSampling is to handle long periods in which the camera wearer is static, hence, the camera is not translating. In these cases, both the fundamental matrix and the FOE estimations can become unstable, leading to wrong cost assignments (low penalty instead of high) to graph edges. The appearance and velocity terms are more robust and help reduce the number of outlier (shaky) frames in the output.

\subsection{Stereo}

\renewcommand{\tabcolsep}{0.1cm}
\begin{table}[t]
    \centering

    \begin{tabular}{lcccc}
        \toprule[1.5pt]
        \specialcell{\bf \tabletext Name}  &\specialcell{\bf \tabletext Resolution} &\specialcell{\bf \tabletext Camera} & \specialcell{\bf \tabletext Input\\ \bf \tabletext Frames} &\specialcell{\bf \tabletext Stereo Pairs \\ \bf \tabletext Extracted} \\ \midrule
        \tabletext Walking1		&	 $1280$x$960$	&	\tabletext Hero2  &	$330$ & $20$	 \\
        \tabletext Walking4		&	 $1920$x$1080$	& Hero3  &	\tabletext $2870$	&  $116$	 	\\
        \tabletext Walking5		&	 $1920$x$1080$	& Hero3  &	\tabletext $1301$  &  	$45$ 	\\
        \bottomrule[1.5pt] \\
    \end{tabular}
    \caption{Sequences used for stereo evaluation. The sequence `Walking1' was shot by \cite{hyperlapse-dataset}. The other two were shot by us.}
    \label{tb:stereo_sequences}
\end{table}

\begin{comment}
\begin{figure}[t]
    %success case
    \centering
    \subfigure{\includegraphics[height=0.2\linewidth]{figures/fig-stereo-walking5-L-frame_00618_small.jpg}}
    \subfigure{\includegraphics[height=0.2\linewidth]{figures/fig-stereo-walking5-R-frame_00622_small.jpg}}
    \subfigure{\includegraphics[height=0.2\linewidth]{figures/fig-stereo-walking5-output.png}}
     \label{fig:stereo_walking5}
    \caption{Stereo Results: Successful case. The proposed approach can compute stereo video from really difficult egocentric video. The first two images show the input frame and third the anaglyph composition. Please use red-cyan anaglyph glasses with zoom level 800\% for best view of third frame. In case glasses are not available, focus on the red shift as an indication of disparity}
\end{figure}
\end{comment}

\begin{figure}[t]
    %failure case
    \centering
    %\subfigure{\includegraphics[height=0.25\linewidth]{figures/fig-stereo-walking1-L-frame_019381_small.jpg}}
    %\subfigure{\includegraphics[height=0.25\linewidth]{figures/fig-stereo-walking1-R-frame_019392_small.jpg}}
    \subfigure{\includegraphics[width=0.45\linewidth]{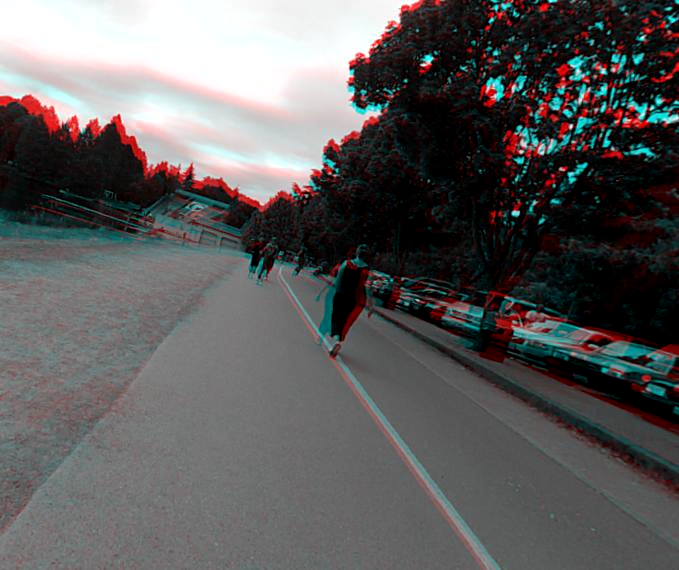}}

    \caption{Stereo failure case. The proposed framework is challenged by the presence of moving objects and registration failures. The disparity perception is presented incorrectly in the example shown because of registration failure. The image shows the anaglyph composition. Best viewed with red-cyan anaglyph glasses at a zoom level of 800\%}
        \label{fig:stereo_walking1}
\end{figure}

Table \ref{tb:stereo_sequences} gives the description of some of the sequences we experimented with for generating stereo video from a monocular egocentric camera. We use publicly available \cite{hyperlapse-dataset} as well as sequences we shot ourselves. Fig.~\ref{fig:stereo_walking4} shows some stereo frames generated by our algorithm.

Registration failure and presence of moving objects pose a significant challenge to the proposed stereo generation framework. Objects present very close to the wearer also disturb the stereo perception. Fig.~\ref{fig:stereo_walking1} shows one such failure instance where the disparity perception has been wrongly computed because of multiple registration failures.

\section{Conclusion}
\label{sec:concl}

We propose a novel frame sampling technique to produce stable fast forward egocentric videos. Instead of the demanding task of $3D$ reconstruction and rendering used by the best existing methods, we rely on simple computation of the epipole or the FOE. The proposed framework is very efficient, which makes it practical for long egocentric videos. Because of its reliance on simple optical flow, the method can potentially handle difficult egocentric videos, where methods requiring $3D$ reconstruction may not be reliable.

We have also presented an approach to use the head motion for generation of stereo pairs. This turns a nuisance into a feature.\\

\noindent\textbf{Acknowledgement:} This research was supported by Intel ICRI-CI, by Israel Ministry of Science, and by Israel Science Foundation.

{\small
\bibliographystyle{ieee}
\bibliography{ego-ff}
}

\end{document}